\documentclass[authoryear,3p]{elsarticle}

\usepackage{pdflscape}
\usepackage{lineno,hyperref}
\usepackage{mathtools}
\usepackage{multirow}
\usepackage{multicol}
\usepackage{booktabs}
\usepackage{listings}
\usepackage[retain-explicit-plus]{siunitx}
\usepackage{xcolor}
\usepackage{footmisc}
\usepackage[ruled,vlined]{algorithm2e}
\modulolinenumbers[5]

\journal{Journal of King Saud University - Computer and Information Sciences}
\biboptions{authoryear}

\begin{document}

\begin{frontmatter}
\lstset{basicstyle=\normalsize\ttfamily,breaklines=true}

\title{Explainable Machine Learning multi-label classification of Spanish legal judgements}

\author[mymainaddress]{Francisco de Arriba-P\'erez}
\ead{farriba@gti.uvigo.es}
\author[mymainaddress]{Silvia Garc\'ia-M\'endez\corref{mycorrespondingauthor}}
\ead{sgarcia@gti.uvigo.es}
\author[mymainaddress]{Francisco J. Gonz\'alez-Casta\~no}
\ead{javier@det.uvigo.es}
\author[mymainaddress]{Jaime González-González}
\ead{jaimegonzalez@gti.uvigo.es}
\address[mymainaddress]{Information Technologies Group, atlanTTic, University of Vigo, EI Telecomunicaci\'on, Campus Lagoas-Marcosende, Vigo, 36310, Spain}

\cortext[mycorrespondingauthor]{Corresponding author: sgarcia@gti.uvigo.es}

\begin{abstract}
Artificial Intelligence techniques such as Machine Learning (\textsc{ml}) have not been exploited to their maximum potential in the legal domain. This has been partially due to the insufficient explanations they provided about their decisions. Automatic expert systems with explanatory capabilities can be specially useful when legal practitioners search jurisprudence to gather contextual knowledge for their cases. Therefore, we propose a hybrid system that applies \textsc{ml} for multi-label classification of judgements (sentences) and visual and natural language descriptions for explanation purposes, boosted by Natural Language Processing techniques and deep legal reasoning to identify the entities, such as the parties, involved. We are not aware of any prior work on automatic multi-label classification of legal judgements also providing natural language explanations to the end-users with comparable overall quality. Our solution achieves over \SI{85}{\percent} micro precision on a labelled data set annotated by legal experts. This endorses its interest to relieve human experts from monotonous labour-intensive legal classification tasks.
\end{abstract}


\end{frontmatter}

\section{Introduction}\label{sec:introduction}

Artificial Intelligence (\textsc{ai}) techniques have not been exploited to their maximum potential in the legal field \citep{Qiu2020} despite the vast amounts of judgements (sentences) that lawyers must review \citep{Zhou2022}. For example, between 2013 and 2016, the number of cases in the Spanish legal system increased by \SI{141}{\percent}\footnote{Available at \url{https://www.poderjudicial.es/cgpj/es/Poder-Judicial/Tribunal-Supremo/Portal-de-Transparencia/Te-puede-interesar---/Estadisticas-}, August 2022.}. Thus, automatic legal data analytics deserve attention to improve its productiveness.

This type of analysis must rely on existing legal databases \citep{Csanyi2022}, which organise judgements by jurisdictions. These databases are helpful as contextual information when legal practitioners such as lawyers, barristers and solicitors look for similar court decisions. However, human-conducted identification tends to be biased by personal points of view and this is exacerbated by the different styles of legal writing and lack of a unified formal structure \citep{Csanyi2022}.

Nowadays, many industrial sectors take advantage of the most recent advances in Natural Language Processing (\textsc{nlp}) techniques and Machine Learning (\textsc{ml}) algorithms \citep{Oussous2018,Tang2020,Roh2021}, and the legal field is no exception, for example for summarisation \citep{Kanapala2019} and anonymisation \citep{DiMartino2021} purposes. Previous work has already addressed multi-label legal document classification \citep{Song2022}. However, there are few freely accessible annotated data sets, their quality is often low, and they tend to be either over- or under-inclusive. This is the case of legal search engines based on keywords \citep{Csanyi2022}.

Particularly, multi-label classification seeks to relate instances to several labels simultaneously. It has been applied to text \citep{Tao2020} and image analysis \citep{Khan2021,Wu2021}, among other problems. Real world applications include sentiment and emotion analysis \citep{Jabreel2019}, medical text classification \citep{Zhang2018,Teng2022} and social network mining \citep{Tao2020}. In addition to more traditional approaches to this type of classification \citep{Aljedani2021}, we can cite one-vs-all \textsc{ml} \citep{Sengupta2021} and deep learning models \citep{Caled2022}.

Multi-label legal classification of judgements, the focus of this work, is a rather natural approach, because each judgement often can be assigned different law categories at a time. Take a case with several charges (embezzlement, blackmail, mugging, etc.) as an example, involving economics and criminal law. Real legal data sets are challenging to many multi-label classification approaches due to the highly imbalanced law categories in them (previous research suggests that dependency modelling may be unnecessary when all labels are sufficiently represented \citep{Burkhardt2018}). This can be helped by deep legal reasoning like prior identification of the parties to generate special features. Accordingly, we propose an in-depth entity detection scheme.

Typically, multi-label classification follows two different strategies: transformation and adaptation \citep{Tarekegn2021}. The former converts a multi-label space problem into a combination of several single-label sub-problems handled by separate binary or multi-class classifiers. This traditional approach is simple but highly effective when sufficient training data are available. Its main drawback is the large number of models needed. Adaptation addresses multi-label classification directly with advanced classifiers.

Table \ref{tab:terms} provides an overview of \textsc{ml} classification approaches according to the number of classes and labels. The problem in this work belongs to the last category, the multi-label multi-class approach.

\begin{table}[!htbp]
\centering
\caption{\label{tab:terms}Classification approaches by number of classes and labels.}
\begin{tabular}{lrr}
\toprule
\textbf{Classification} & \textbf{\# Classes} & \textbf{\# Labels} \\\midrule
Binary & 2 & 1\\
Multi-class & \textgreater{}2 & 1\\
Multi-label (binary) & 2 & \textgreater{}1\\
\textbf{Multi-label multi-class} & \textgreater{}2 & \textgreater{}1\\
\bottomrule
\end{tabular}
\end{table}

Explainability can be defined as the relation between the actual and the expected behaviours of a system. It must be addressed along with other closely related relevant aspects \citep{Vilone2021}: comprehensibility (quality of the method and language used), informativeness (to provide useful knowledge to end-users), transparency (the capacity to describe expected or unexpected behaviour of the systems) and understandability (the capacity to make the models comprehensible). 

We are interested in providing natural language explanations to end-users about supervised multi-label multi-class classifications of legal judgements (note that unsupervised \textsc{ml} models usually exhibit low explainability because their outputs are incomprehensible to human operators \citep{Canhoto2021}).

The rest of this work is organised as follows. Section \ref{sec:related_work} reviews related work on legal classification taking into account both multi-label approaches and explainability. Section \ref{sec:system} presents our classification problem and the architecture of our solution. Section \ref{sec:results} provides details about the experimental data set and discusses the results to validate our approach. Finally, Section \ref{sec:conclusions} concludes the paper.

\section{Related work}
\label{sec:related_work}

As previously said, legal classification can be performed at binary (by choosing between two complementary sets of categories) \citep{Medvedeva2020}, multi-class (by selecting among several categories) \citep{Bambroo2021,El-Alami2021} or multi-label level (by associating each instance to multiple categories) \citep{Song2022}. 

Next, multi-label algorithms can be divided into three types \citep{Zhang2014} depending on their treatment of label dependency and correlation: first-order dependency agnostic solutions (\textit{i.e.}, when labels are treated independently); second-order solutions that consider pairwise label dependencies between relevant and non-relevant labels through rankings; and high-order architectures modelling label relations, such as classifier chains \citep{Teisseyre2021}. The latter scale poorly and are computationally expensive compared to the first two types.

The most straightforward transformation of multi-label problems simply implements a binary classifier per target class. Another alternative is to train a classifier on all existing combinations of classes in the label sets of the individuals (without repetition) \citep{Qiu2020}. Some authors employ intermediate output knowledge as input data to subsequent classification algorithms across a classifier chain, or ensemble scheme \citep{Teisseyre2021}. \cite{Aljedani2021} proposed a hierarchical multi-label \textsc{ml} classification solution for \cite{Burkhardt2018} used supervised greedy layer-wise training to learn labels and then represented their dependencies.

Adaptive approaches to multi-label \textsc{ml} classification enhance existing algorithms or are completely new proposals, such as the multi-label k-Nearest Neighbour (\textsc{mlknn}) algorithm \citep{Skryjomski2019} and the Rank-\textsc{svm} \citep{Wu2020}. 

Transformation and adaptation approaches can rely on diverse general strategies, such as one-vs-all \citep{Sengupta2021}, tree ensembles \citep{Moyano2018}, embedding solutions \citep{Caled2022,Caled2019} and deep learning \citep{Caled2022}. The training and testing processes of the one-vs-all strategy are computationally consuming, however. Tree ensembles recursively divide the multi-label space using a Decision Tree \citep{Trabelsi2019} or a similar algorithm with simple classifiers at the nodes for prediction purposes. Embedding strategies consider label correlation and try to reduce the multi-label space by using label vectors in the training stage and decomposing them when predicting new samples \citep{Chen2016}. The output layer of a deep learning network can be adapted to multi-label operation with sigmoid functions \citep{Liu2021} and attention mechanisms \citep{Song2022}.

The focus of this work is the explainable classification of legal judgements, a type of legal texts with distinctive structure, style and vocabulary that conform the jurisprudence. \cite{Bambroo2021} performed a pure multi-class classification of \num{300000} U.S. judgement documents, that is, each with a single label. This was also the case of the comparison of traditional \textsc{ml} classification solutions with deep learning models by \cite{Chen2022}. They reported that the Random Forest \citep{Parmar2019} algorithm outperformed deep learning models using domain-concept-based features on a data set of \num{30000} U.S. judgements of \num{50} different categories. \cite{Csanyi2022} decomposed the multi-label classification of judgements into binary classification sub-problems with a transformation approach requiring over a hundred classifiers. \cite{Prajapati2021} proposed the K-way Tree based eXtreme Multi-Label Classifier to maintain feature correlations using a clustering algorithm. This approach was tested over the legal Eurlex\footnote{Available at \url{https://data.europa.eu/data/datasets/eur-lex-statistics?locale=en}, August 2022.} data set with almost \num{20000} instances, among other general-purpose data sets. Finally, \cite{Song2022} analysed the \textsc{posture50k} annotated data set of \num{50000} judgements. They applied domain-specific pre-training and label-attention mechanisms based on the RoBERTa model for extreme multi-label legal classification.

We remark that there exists research on the classification of other types of legal documents outside the scope of this work. Both \cite{Sengupta2021} and \cite{Caled2022,Caled2019} focused on legislative texts. These are substantially different to court judgements and require specific methodologies.

Explainability is a recent topic of interest in expert systems, driven by new technological advances and the reluctance of end users to trust the decisions by opaque \textsc{ai} solutions \citep{Branting2017,Atkinson2020,BarredoArrieta2020}. Among the variety of explanatory techniques in the literature (such as graph-based \citep{Cai2022}, model agnostic \citep{Leal:2022}, word embeddings \citep{Qureshi:2019} and visual explanations \citep{Carvalho2019,Ye:2021}), we pay special attention to natural language representation of new knowledge through graphic visualisations and templates to foster not only accurate but also unbiased and trustworthy decision support to the practitioners. 

The use of \textsc{ml} models was limited in this context until very recently, partially owing to their unsatisfactory explanatory capabilities. Consequently, \textsc{ml} models can be divided into transparent and opaque models. The former are self-interpretable or explainable through external modules (\textit{e.g.}, decision trees, rules or regression), and the latter behave as black boxes (\textit{e.g.}, artificial neural networks). It has been stated that transparent, explainable and interpretable models contribute to responsible \textsc{ml} \citep{BarredoArrieta2020}.

\cite{Branting2021} presented a system to explain human annotations of legal documents, that is, they did not classify these documents automatically, and therefore their approach lies outside the scope of our work. \cite{Caled2022,Caled2019} only interpreted their model using attention weights, \textit{i.e.}, without natural language descriptions, and, as previously said, their problem is different. Therefore, our work contributes to the state of the art in legal judgement classification with a hybrid system that combines \textsc{ml} for prediction with visual and natural language representation down to feature level for explanation (we describe both the result of the process and the process itself). Judgement classification can also benefit from deep legal reasoning to generate relevant features \citep{Thomas2021}. Accordingly, we include in-depth entity detection to extract relevant judicial entities such as case type, court and jurisdiction, among others. As recapitulated in Section \ref{sec:literature_comparison}, no relevant prior competing works considered the automatic explanation of the classification of judgements, so they did not take explanation into account in their selections of methodologies. Besides, either the data sets were too small, the training times were far superior to ours or the performance results were worse.

\subsection{Research contribution}
\label{sec:contribution}

Table \ref{tab:related_work} shows an overview of the systems in the literature for multi-label classification of legal judgements taking into account language, size of the experimental data sets, techniques and explainability. In light of this, our proposal is the first multi-label classification system with advanced visual and natural language explanation capabilities with competitive performance (see Section \ref{sec:literature_comparison} for a detailed experimental comparison of our system with related competing solutions in the literature).

\begin{table}[!htbp]
 \centering
\caption{\label{tab:related_work}Multi-label classification of legal judgements (\textsc{cnn}: Convolutional Neural Network, \textsc{nl}: Natural Language).}
 \begin{tabular}{lcccc}
 \toprule
 \textbf{Work} & \textbf{Language} & \textbf{Data set size (documents)} & \textbf{Technique} & \textbf{Explainability} \\
 \midrule
 
 \cite{Qiu2020} & Chinese & \num{143} & \textsc{cnn} & No\\

 \cite{Csanyi2022} & Hungarian & \num{173892} & Supervised \textsc{ml} & No\\

 \cite{Song2022} & English & \num{50000} & Deep learning & No\\\midrule

 \textbf{Our proposal} & Spanish & \num{106806} & Supervised \textsc{ml} & Visual \& \textsc{nl}\\

\bottomrule
 \end{tabular}
\end{table}

Adequate preprocessing and feature engineering of unstructured judgements, as well as \textsc{nlp} techniques, allow us to take full advantage of \textsc{ml} algorithms. The good performance of our solution endorses its suitability to relieve human experts from monotonous and labour-intensive legal classification tasks.

\section{System architecture}
\label{sec:system}

\begin{figure}[!htbp]
 \centering
 \includegraphics[scale=0.20]{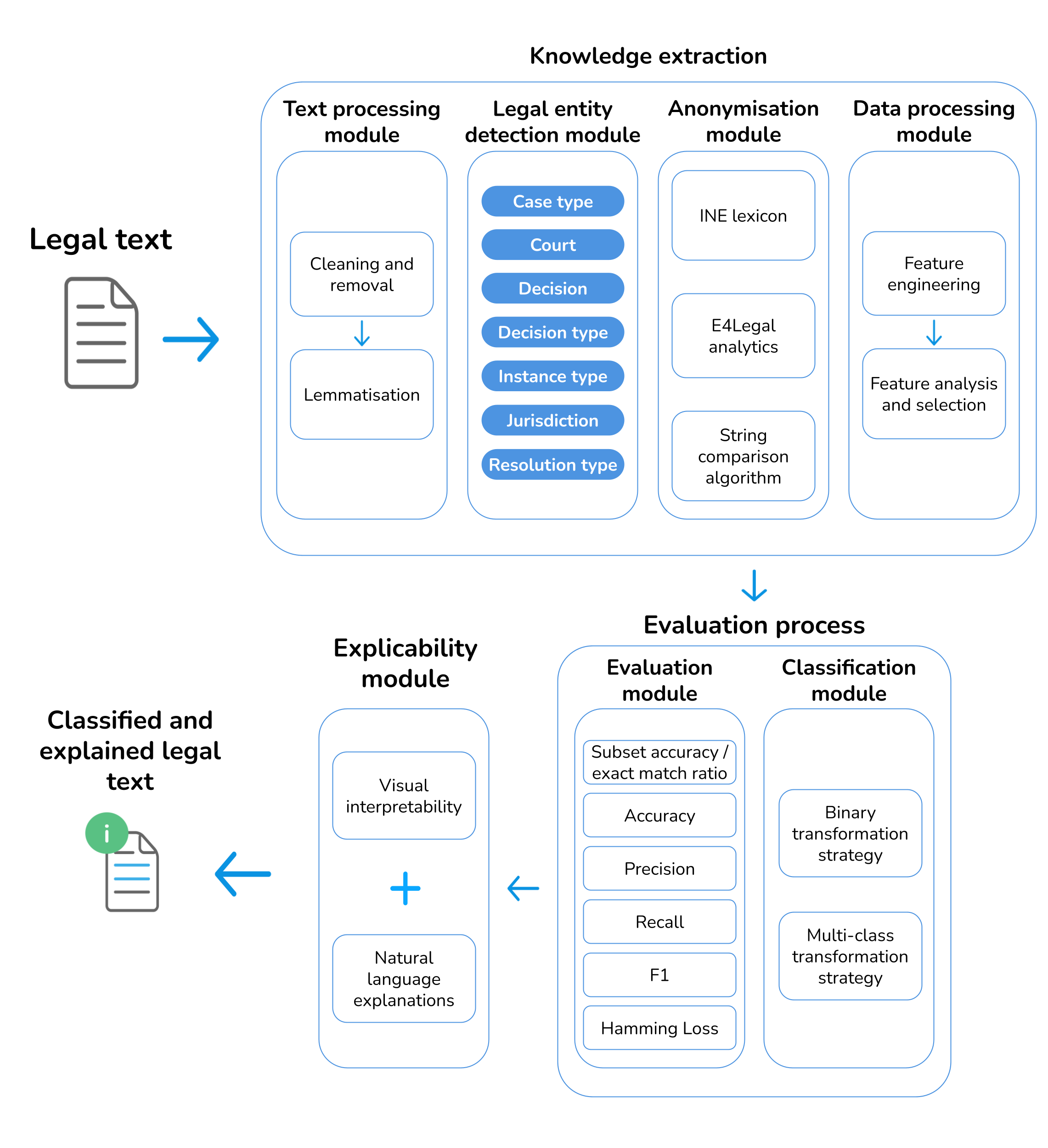}
 \caption{System scheme.}
 \label{fig:architecture}
\end{figure}

In this section, we describe in detail our approach for multi-label multi-class classification of legal judgements. Specifically, the system must predict at most three labels for each legal document, and each class of those labels is composed of four elements: a \textit{substantive order} (penal, civil, social, administrative, civil/mercantile, mercantile or tributary) plus three \textit{law categories} (see Section \ref{sec:dataset}). Figure \ref{fig:architecture} shows the system scheme. It is composed of (\textit{i}) a text processing module for text cleaning and lemmatisation (Section \ref{sec:text_processing}), (\textit{ii}) a legal entity detection module (Section \ref{sec:judicial_module}) to automatically extract from a judgement text the case, instance and resolution types, the court, the decision and its type, and the jurisdiction; (\textit{iii}) an anonymisation module (Section \ref{sec:anonymisation_module}), (\textit{iv}) a data processing module (Section \ref{sec:dataprocessing_module}) composed of feature engineering (Section \ref{sec:feature_engineering_architecture}), analysis and selection (Section \ref{sec:feature_selection_architecture}); (\textit{v}) a classification module (Section \ref{sec:classification_architecture}) with binary and multi-class transformation strategies, (\textit{vi}) an evaluation module (Section \ref{sec:evaluation_metrics}) using standard multi-label metrics and (\textit{vii}) an explainability module (Section \ref{sec:explainability_architecture}) that produces visual and natural language explanations.

\subsection{Text processing module}
\label{sec:text_processing}

This first module removes unnecessary elements and reduces the word space for the \textsc{ml} models.

\begin{itemize}
    \item \textbf{Cleaning and removal}. Page breaks, tab stops, \textsc{ulr}s, punctuation marks and redundant spaces are removed from the legal judgements. Stop-words are also eliminated.
    \item \textbf{Lemmatisation}. The textual content that is extracted from the judgement is firstly tokenised into words and then these elements are lemmatised.
\end{itemize}

\subsection{Legal entity detection module}
\label{sec:judicial_module}

A judgement text of the Spanish legal system is usually divided into a \textit{heading} section describing the type of legal resolution, the identifiers and the parties involved in the case; \textit{pleas of fact} detailing the facts under discussion and their previous story; \textit{pleas of law} relating the case to current laws and legal precedents; and the \textit{court decision}. Moreover, each legal judgement usually includes a General Identification Number\footnote{This identification number is assigned where the legal procedure starts and is kept even if the court changes.} that indicates the province (digits 1-5), court (digits 6-7), jurisdiction (digit 8), year (digits 9-12), and sentence identifier (digits 13-19).

Next we describe the extraction of relevant information to produce special features using lexica revised by legal experts. 

\begin{itemize}

\item \textbf{Case type}. We use a publicly available official lexicon\footnote{Available at \url{https://bit.ly/3yO8u5h}, August 2022.} and the regular expression $\backslash d+\backslash d\{1,4\}$ to detect the case type among the possible types in Spanish jurisprudence. Official abbreviations are also considered. In case of multiple matches, the first match is selected.

\item \textbf{Court}. We rely on a lexicon 
that is extracted from 
 the General Access Point of the Spanish Court Administration\footnote{Available at \url{https://bit.ly/3wCOEbO}, August 2022.} to detect the court name with a regular expression.

\item \textbf{Decision}. The text in the \textit{court decision} section is used to extract the keywords to be checked against the decision lexicon\footnote{Available at \url{https://bit.ly/3Ntle5r}, August 2022.}. When there exist several keywords, the decision is considered a \textit{multiple decision}. Otherwise, the decision is the detected keyword itself.

\item \textbf{Decision type}. It depends on the legal document. We consider \textit{substantive} and \textit{procedural} decision types. The first applies to judgements (sentences). The second to orders and decrees.

\item \textbf{Instance type}. If the case type contains ``recurso''; ``apelación''/ ``suplicación''; or ``casación''/ ``unificación'', the instance type is \textit{first}, \textit{second} or \textit{third}, respectively. Otherwise, it is considered a \textit{higher} instance of the Spanish legal system.

\item \textbf{Jurisdiction}. There exist three different ways to extract the jurisdiction (which can be \textit{civil}, \textit{contentious-administrative}, \textit{penal} or \textit{social}) from Spanish legal documents, in decreasing order of effectiveness according to our experience. First, we extract the judicial division (it appears next to the court information) in charge of the case, which most of the time includes directly the type of court, that is, the jurisdiction. Another way consists in identifying the General Identification Number of the legal judgement, of which, as previously said, the eighth digit refers explicitly to the jurisdiction. Finally, it can be extracted from the case type lexica\footnote{Available at \url{https://bit.ly/3GnIlMJ}, August 2022.}. 

\item \textbf{Resolution type}. It can be found at the end of the \textit{heading} section just before the \textit{pleas of fact}. It is detected using a regular expression containing the three possible values, ``sentencia'' (\textit{judgement/sentence}), ``orden'' (\textit{order}) and ``decreto'' (\textit{decree}).

\end{itemize}

\begin{figure*}[!htbp]
 \centering
 \includegraphics[scale=0.2]{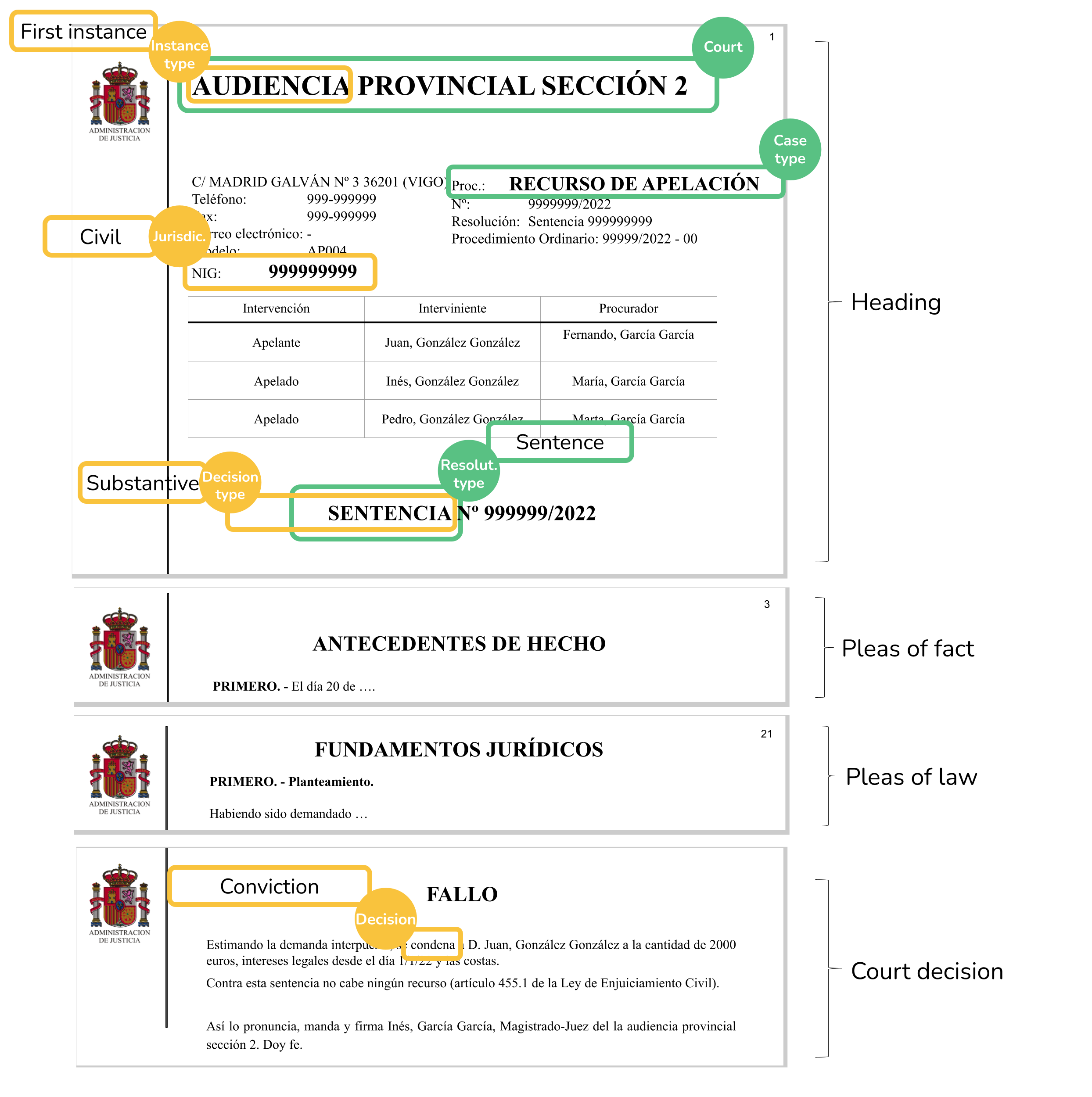}
 \caption{Structure of a legal sentence and identification of relevant entities.}
 \label{fig:sentence}
\end{figure*}

Figure \ref{fig:sentence} shows an example taken from a real judgement of the Spanish legal system.

\subsection{Anonymisation module}
\label{sec:anonymisation_module}

It is essential to remove proper names to avoid bias in the \textsc{ml} models. The anonymisation proceeds as follows (we indicate all the resources we use in footnotes):

\begin{enumerate}
    \item \textbf{Detection of references to people}:
    \begin{enumerate}
        \item Personal titles\footnote{Available at \url{https://bit.ly/39KWlDG}, August 2022.}
        \item Implicit references\footnote{Available at \url{https://bit.ly/3GbDvBZ}, August 2022.}
        \item Corporate legal forms\footnote{Available at \url{https://bit.ly/3Nssb6T}, August 2022.}
    \end{enumerate}
    \item \textbf{Replacement}. The previous references are replaced by tags \texttt{@Attorney}, \texttt{@Corporate}, \texttt{@Judge}, \texttt{@Lawyer} and \texttt{@Person} as required. Next, we identify the first and last names next to the tags, using data from the Spanish Statistics National Institute\footnote{Available at \url{https://www.ine.es/dyngs/INEbase/es/operacion.htm?c=Estadistica_C&cid=1254736177009&menu=ultiDatos&idp=1254734710990}, August 2022.}. Finally, elastic indexing systems provided by E4Legal Analytics \textsc{sl}\footnote{Available at \url{https://www.emerita.legal}, August 2022.} through Solr\footnote{Available at \url{https://solr.apache.org}, August 2022.} allow us to verify and correct both the people and companies that are detected and their roles in the legal documents (by excluding \texttt{@Person}). A string comparison algorithm detects similar elements (see Section \ref{sec:anonymisation_results}).
\end{enumerate}

\subsection{Data processing module}
\label{sec:dataprocessing_module}

It is composed of two stages: (\textit{i}) feature engineering and (\textit{ii}) feature analysis and selection. This module improves the effectiveness of the \textsc{ml} models for multi-label legal classification.

\subsubsection{Feature engineering}
\label{sec:feature_engineering_architecture}

This step extracts valuable knowledge from the textual content 
 to produce the features in Table \ref{tab:features}:
 
\begin{itemize}
 \item \textbf{Textual features}. $N$-grams.
 
 \item \textbf{Categorical features}. As previously mentioned, \textit{case type}, \textit{court}, \textit{decision}, \textit{decision type}, \textit{instance type}, \textit{jurisdiction} and \textit{resolution type} (see Section \ref{sec:judicial_module}).
\end{itemize}

\begin{table*}[!htbp]
\centering
\caption{\label{tab:features} Features and target of the \textsc{ml} models.}
\begin{tabular}{llll}
\toprule
\bf Type & \bf Id. & \bf Feature name & \bf Description \\\midrule

Textual & 1 & $N$-grams & \begin{tabular}[c]{@{}p{8.5cm}@{}} $N$-grams.\end{tabular}\\
 
 \midrule

\multirow{6}{*}{Categorical} & 2 & Case type & \begin{tabular}[c]{@{}p{8.5cm}@{}} Case type within Spanish jurisprudence. \end{tabular}\\

 & 3 & Court & \begin{tabular}[c]{@{}p{8.5cm}@{}} 
 Court in charge of the case.\end{tabular}\\
 
 & 4 & Decision & \begin{tabular}[c]{@{}p{8.5cm}@{}} 
 Decision in the legal judgement or \textit{multiple decision}.\end{tabular}\\
 
 & 5 & Decision type & \begin{tabular}[c]{@{}p{8.5cm}@{}} 
 \textit{Substantive} or \textit{procedural} decision type.\end{tabular}\\
 
 & 6 & Instance type & \begin{tabular}[c]{@{}p{8.5cm}@{}} 
 Instance in charge of the case: first, second, third or higher level.\end{tabular}\\
 
 & 7 & Jurisdiction & \begin{tabular}[c]{@{}p{8.5cm}@{}} 
 Type of court in charge of the case.\end{tabular}\\
 
 & 8 & Resolution type & \begin{tabular}[c]{@{}p{8.5cm}@{}} 
 \textit{Judgement/sentence}, \textit{order} or \textit{decree}.\end{tabular}\\
 
\bottomrule
\end{tabular}
\end{table*}

\subsubsection{Feature analysis and selection}
\label{sec:feature_selection_architecture}

The target to predict per legal judgement is a set with up to three labels, each of which takes class values consisting of a substantive order of the legal judgement plus three choices of law categories within that order.
We analyse the relation between the target and the features in Table \ref{tab:features} using Spearman's rank correlation coefficient \eqref{spearman} \citep{DeWinter2016}. This measures the statistically monotonic (linear or not) relationships between continuous and discrete features, particularly between the target and the rankings of the independent features in an equispaced 10-step discretised scale in inverse order of magnitude. Values range from -1 to +1, where the extremes indicate perfectly monotonic relations.

\begin{equation}
r_s = \frac{cov(R(X),R(Y))}{\sigma_{R(X)}\sigma_{R(Y)}}
\label{spearman}
\end{equation}

\noindent
Here, $cov(R(X), R(Y))$ is the covariance of the rank variables and $\sigma_{R(X)}$ $\sigma_{R(Y)}$ are the standard deviations of the rank variables. The system selects the features with a Spearman's rank correlation coefficient above a configurable threshold, as well as the significant features at the training stage, based on the importance weights delivered by an algorithm-dependent meta-transformer.

\subsection{Classification module}
\label{sec:classification_architecture}

The strategies we follow to predict the target are:

\begin{itemize}
    \item \textbf{Binary transformation strategy} (\textsc{bts}). This strategy transforms the multi-label multi-class problem into a multi-label binary problem. In other words, the multi-label space is transformed into the combination of several binary single-class sub-problems by using separate classifiers. A binary matrix is defined with as many columns as classes and as many rows as individuals in the training set, and a binary classifier is trained per column. As a result, a Boolean vector per entry indicates the classes predicted at the output. For a given data set of $n$ individuals $\{x_1,\ldots,x_n\}$ with respective label sets $\{L_1,\ldots,L_n\}$, $|L_i|=l_i$, $i=1,\ldots,n$ (in this work $l_i \in [1,3]$), let $C$ be the set of possible classes, $|C|=m$. That is, $\forall \lambda \in C$ $\rightarrow$ $\lambda \in \cup_{j=1}^n L_{j} $; and $\forall \lambda_i, \lambda_j \in C \mid i \neq j \rightarrow \lambda_i \neq \lambda_j$. Then, the Boolean labels of the $j$-th \textsc{bts} binary classifier, $j=1,\ldots,m$, for the $i$-th individual, can be defined as:
    
    \begin{equation}
        \beta_{ij} = \left\{ \begin{array}{lcc}
       0 & \mbox{if} & \lambda_j \notin L_i\\
       1 & \mbox{if} & \lambda_j \in L_i \\
       \end{array} \right.
    \end{equation}
    
    \item \textbf{Multi-class transformation strategy} (\textsc{mts}). This strategy transforms the multi-label multi-class problem into a plain multi-class problem. In it, the label sets of the individuals are considered combinations of label classes, each of which becomes a class of the multi-class problem (in other words, if the elements of a label set are a permutation of the elements of another label set, both label sets correspond to the same class in the transformed problem). Specifically, let $C_M=\{L'_1,\ldots,L'_p\}$ be the set of all different combinations of classes in the label sets, i.e. $\forall L'_i, L'_j \in C_M \mid i \neq j \rightarrow L'_i \neq L'_j$, and combination $L'_i$ corresponds to label sets $L_j$ and $L_k$, $j,k \in \{1,\ldots,n\}$, if $| L'_i|$=$| L_j|$=$| L_k |$; and $\forall \lambda \in L_j \cup L_k$ $\rightarrow$ $\lambda \in L'_i$. Let us define $f_r:L\rightarrow L'$ as the reordering function that transforms a label set $L$ into the corresponding combination of label classes $L'$ so defined. 
    Then, the multiclass transformation of the label set of the $i-th$ individual produces an integer label $\alpha_i$ for that individual \eqref{mtslabel}. Logically, in case the classes are string characters, as in this work, label set reordering can simply consist in ordering the labels alphabetically.
    
    \begin{equation}
    \label{mtslabel}
        \alpha_i = j \; \mbox{if} \; f_r(L_i)=L'_j, \; \alpha_i \in [1,p]
    \end{equation}

\end{itemize}

To shed light on the two strategies used, Listing \ref{cendoj} shows a sample of a legal judgement as manually annotated\footnote{Available at \url{https://www.poderjudicial.es/search/AN/openDocument/fae66aca1afeb8e2a0a8778d75e36f0d/20220920}, September 2022.}, referred to as File $i$ in the sequel. Tables \ref{tab:bts_cendoj} and \ref{tab:mts_cendoj} provide the corresponding \textsc{bts} and \textsc{mts} labels, respectively. \textsc{bts} labelling of a legal judgement represents each annotated combination of a substantive order and three law categories in a binary space (Table \ref{tab:bts_cendoj}). \textsc{mts} labelling groups all the possible annotations in a single categorical label (Table \ref{tab:mts_cendoj}).

\begin{lstlisting}[frame=single,caption={Legal judgement with two manual annotations regarding substantive order and law categories.}, label={cendoj}]
Substantive order: civil/mercantile
Law categories: real rights, guarantee real rights, mortgage law
Excerpt: Versa el juicio sobre la validez o no (y en su caso sus consecuencias) de dos de las cláusulas (suelo -incluido el acuerdo de su eliminación- e interés de demora) de la escritura de adjudicación de vivienda de protección oficial y subrogación modificativa con ampliación en préstamo hipotecario de fecha XXXXXX autorizada por el notario de XXXXXX con el número XXXXXX de su protocolo en la que (además de la agrupación de viviendas transmitente) intervinieron quienes son parte en el procedimiento que aquí se resuelve.

Substantive order: mercantile
Law categories: obligations - contracts law, banking - financial market law, banking law
Excerpt: Condeno XXXXXX, XXXX recalcular las cuotas satisfechas aplicando, sin el suelo ni tipo fijo del acuerdo privado, el tipo de interés pactado en la escritura que estuviera vigente en la fecha de devengo de cada cuota (Euribor 12 meses + diferencial - bonificaciones en su caso); de cuyo re/cálculo se dará traslado al actor que podrá presentar liquidación contradictoria; en tal caso el juzgado fijará la cantidad correcta (2) restituir al actor la diferencia entre las cuotas abonadas con aplicación del suelo y tipo fijo del acuerdo privado y las recalculadas sin aplicación de dicha cláusula y pacto, (3) abonar al actor, sobre el importe cobrado en exceso en cada cuota, intereses al tipo legal del dinero desde la fecha de abono de la misma hasta sentencia, e incrementado el tipo en dos puntos desde la fecha de esta sentencia hasta el completo pago.
\end{lstlisting}

\begin{table}[!ht]
\caption{Example of \textsc{bts} labelling from annotated legal judgement.~\label{tab:bts_cendoj}}
\begin{tabular}{cccccc}
\toprule
& \multicolumn{5}{c}{\begin{tabular}[c]{@{}c@{}}\bf Substantive order\\ \bf First law\\ \bf Second law\\ \bf Third law\end{tabular}}\\
\cmidrule{2-6}
& ... & \begin{tabular}[c]{@{}l@{}}Civil/Mercantile\\ Real rights\\ Guarantee real rights\\ Mortgage law\end{tabular} & \begin{tabular}[c]{@{}l@{}}Mercantile\\ Obligations-contracts law\\ Banking-financial market law\\ Banking law\end{tabular} & \begin{tabular}[c]{@{}l@{}}Civil/Mercantile\\ Obligations-contracts law\\ Damages law\\ Road traffic law\end{tabular} & ... \\\midrule
\multicolumn{1}{l}{\bf File $i$} & ... & 1 & 1 & 0 & ...\\
\bottomrule
\end{tabular}
\end{table}

\begin{table}[!ht]
\centering
\caption{Example of \textsc{mts} labelling from annotated legal judgement.~\label{tab:mts_cendoj}}
\begin{tabular}{cl}
\toprule
& \multicolumn{1}{c}{\begin{tabular}[c]{@{}c@{}}\bf Substantive order\\ \bf First law\\ \bf Second law\\ \bf Third law\end{tabular}}\\
\midrule
\bf File $i$ & Civil/Mercantile\\
& Real rights\\
& Guarantee real rights\\
& Mortgage law\\
& \\
& Mercantile\\
& Obligation-contracts law\\
& Banking-financial market law\\
& Banking law\\
&\\

\bottomrule
\end{tabular}
\end{table}

\subsection{Evaluation module}
\label{sec:evaluation_metrics}

Multi-label classification evaluation is complex because prediction, as well as labelling itself, may only be partially correct. Therefore, a careful design of evaluation metrics is necessary to assess the performance \citep{Pereira2018}.

More in detail, we follow the example-based evaluation approach to capture the notions of fully-correct (exact match ratio) and partially-correct predictions (accuracy, precision, recall and Hamming Loss). Next, we list all metrics and their definitions for the sake of clarity:

\begin{itemize}

\item \textbf{Exact match ratio}. This metric considers that partially-correct predictions are incorrect. It is computed as the accuracy of a single-label scenario, where $I$ is the indicator function, $n$ the number of individuals in the test set, $L_i$ the set of annotated labels of individual $i$ and $Z_i$ the set of predicted labels of individual $i$.

\begin{equation}
    \frac{1}{n} \sum_{i=1}^{n}I(L_i=Z_i)
\end{equation}

\item \textbf{Accuracy} per individual is defined as the proportion of correctly predicted labels over the amount of predicted and annotated labels for that individual. Overall accuracy is calculated as the average accuracy across all instances.

\begin{equation}
    \frac{1}{n} \sum_{i=1}^{n}\frac{|L_i \cap Z_i|}{|L_i \cup Z_i|}
\end{equation}

\item \textbf{Precision} per individual is defined as the proportion of correctly predicted labels over the amount of annotated labels. Overall precision is calculated as the average precision across all instances.

\begin{equation}
    \frac{1}{n} \sum_{i=1}^{n}\frac{|L_i \cap Z_i|}{|Z_i|}
\end{equation}

\item \textbf{Recall} per individual is defined as the proportion of correctly predicted labels over the amount of predicted labels. The overall recall is calculated as the average recall across all instances.

\begin{equation}
    \frac{1}{n} \sum_{i=1}^{n}\frac{|L_i \cap Z_i|}{|L_i|}
\end{equation}

\item The \textbf{Hamming Loss} (\textsc{hl}) considers incorrectly predicted labels (prediction error) and relevant non-predicted labels (missing error), normalised over the maximum possible number of label instances in the test set. The closer to 0 (no error), the better.

\begin{equation}
    \frac{1}{|C| \cdot n} \sum_{i=1}^{n} \sum_{\lambda \in C}[I(\lambda \in Z_i \wedge \lambda \notin L_i) + I(\lambda \notin Z_i \wedge \lambda \in L_i)]
\end{equation}

\end{itemize}

Note that for the classification precision, recall and \textit{F}-measure metrics, micro-averaging and macro-averaging are calculated. Firstly, the macro-average approach computes the metric average for each class independently, by treating all classes equally, while the micro-average approach computes the metric average by aggregating the contributions of all classes. These averaging approaches are specially suitable when the experimental data set is class-imbalanced \citep{TaoLiu2004,Liu2009}.

\subsection{Explainability module}
\label{sec:explainability_architecture}

Our multi-label classification system relies on interpretable \textsc{ml} models. These are models that allow extracting visual and natural language explanations at a reasonable cost to describe the classification outcomes, \textit{i.e.}, each combination of substantive order and law categories that is predicted for a certain legal judgement. For example, in the case of tree decision algorithms, as detailed in Section \ref{sec:explainability_results}, the information in relevant decision paths in a visual graph can be exploited to create natural language templates to describe those paths. It is also possible to include the relevance of the different features  involved in a decision as part of its description. These explanations contribute to the transparency of the classifier making the results more acceptable to end users.

\section{Evaluation}
\label{sec:results}

All the experiments were executed on a computer with the following specifications:
\begin{itemize}
 \item Operating System: Ubuntu 18.04.2 LTS 64 bits
 \item Processor: Intel\@Core i9-10900K 2.80 GHz 
 \item RAM: 96GB DDR4 
 \item Disk: 480 GB NVME + 500 GB SSD
\end{itemize}

\subsection{Experimental data set}
\label{sec:dataset}

\begin{figure}[!htbp]
 \centering
 \includegraphics[scale=0.095]{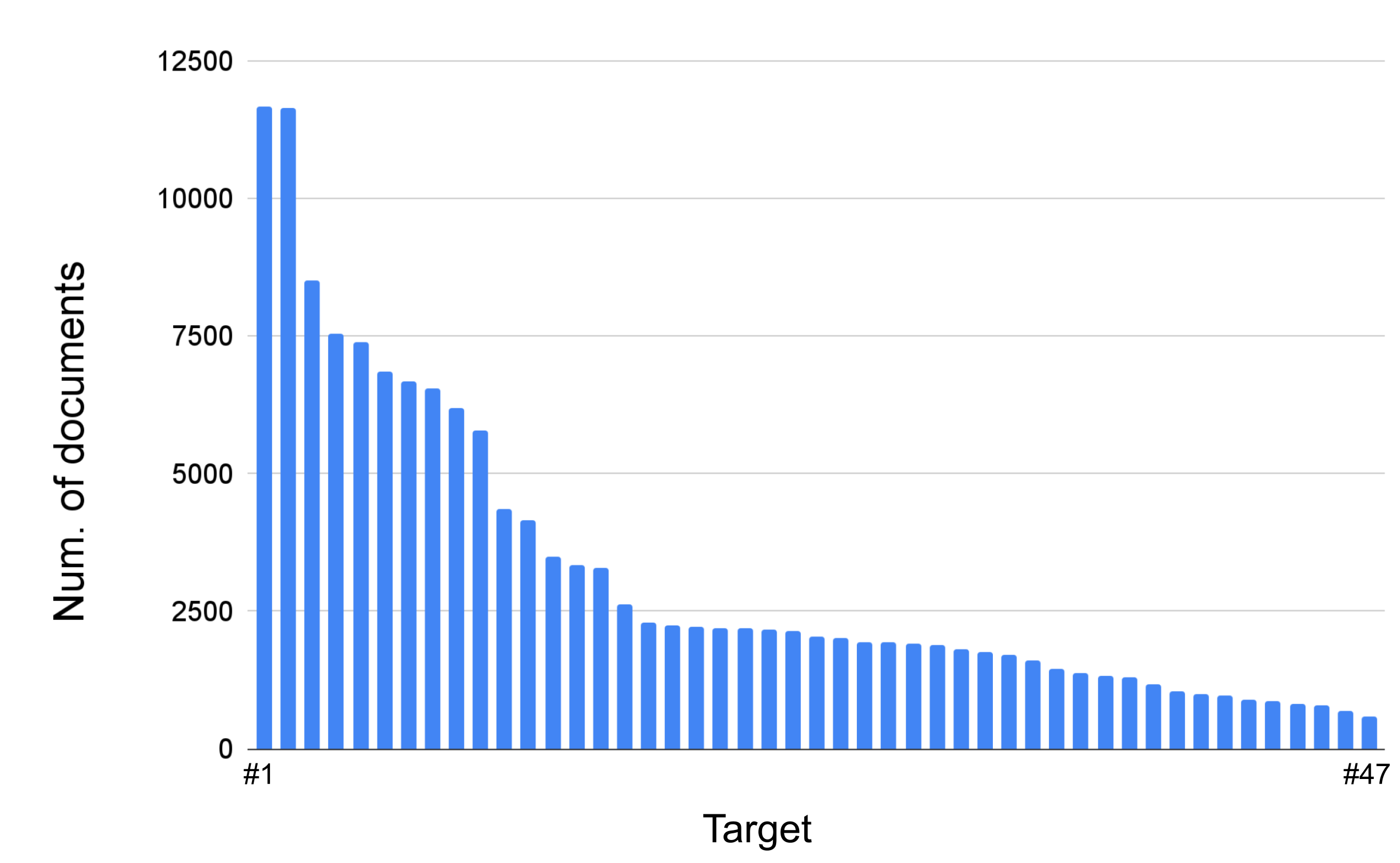}
 \caption{Histogram of targets in the multi-class transformed problem.}
 \label{fig:histogram}
\end{figure}

The experimental data set is composed of \num{106806} judgements that were annotated by lawyers of E4Legal Analytics \textsc{sl}. Each document is annotated with three labels at most, each of these labels consisting of a substantive order and three law categories (see Figure \ref{fig:partial_tree} for examples of classes). There exist \num{47} different classes (targets) in the multi-class transformed problem. Figure \ref{fig:histogram} shows the histogram of documents per transformed classes in the experimental data set. Table \ref{tab:dataset_distribution} shows the distribution of sizes of the label sets. The average label cardinality of the data set (\num{1.39}) is considered high in the literature \citep{Chen2016}.

\begin{table}[!htbp]
\centering
\caption{Distribution of label set sizes in the experimental data set.}
\begin{tabular}{cc}
\toprule \textbf{Label set size} & \textbf{Num. of documents} \\ \midrule
1 & \num{72182} \\
2 & \num{27614} \\
3 & \num{7010} \\
\midrule
Total & \num{106806} \\ \bottomrule
\end{tabular}
\label{tab:dataset_distribution}
\end{table}

\begin{table}[!htbp]
\centering
\caption{\label{tab:substantive_order_distribution}Distribution of documents by substantive order.}
\begin{tabular}{lr}
\toprule
\bf Substantive order & \bf Num. of documents \\
\midrule
Penal             & 57444             \\
Civil             & 28784             \\
Social            & 17259             \\
Administrative    & 15568             \\
Civil/mercantile  & 12965             \\
Mercantile        & 12274             \\
Tributary         & 4146              \\
\bottomrule
\end{tabular}
\end{table}

Table \ref{tab:substantive_order_distribution} shows the distribution of legal judgements in the experimental data set according to their substantive order. 

\subsection{Text processing module}
\label{sec:text_processing_results}

The text processing in Section \ref{sec:text_processing} is implemented as follows: for the lemmatisation we use the spaCy \textsc{nlp} pipeline\footnote{Available at \url{https://spacy.io}, August 2022.} with the \texttt{es\_core\_news\_md}\footnote{Available at \url{https://spacy.io/models/es}, August 2022.} model for Spanish. Stop-words are detected with the \textsc{nltk} toolkit\footnote{Available at\url{https://www.nltk.org}, August 2022.}.

\subsection{Legal entity detection module}
\label{sec:judicial_results}

As described in Section \ref{sec:judicial_module}, we apply regular expressions and legal lexica along with a fast indexing database with Solr technology to generate additional relevant categorical features (see Table \ref{tab:features}).

\subsection{Anonymisation module}
\label{sec:anonymisation_results}

Regarding the anonymisation in Section \ref{sec:anonymisation_module}, we used the jellyfish Python string comparison library\footnote{Available at \url{https://pypi.org/project/jellyfish}, August 2022.} with Jaro distance to detect and unify similar occurrences of first and second names.

As a result, all \num{106806} documents contained sensitive data, and the anonymisation module replaced \num{188710} lawyers' instances, \num{128260} solicitors' instances, \num{266839} judges' instances and \num{22365} proper names of people involved in the cases.

\subsection{Data processing module}
\label{sec:dataprocessing_results}

In this section we present the results of the feature engineering process and the analysis and selection process.

\subsubsection{Feature engineering}
\label{sec:feature_engineering_results}

To produce the textual features (type 1) in Table \ref{tab:features}, we employed \texttt{CountVectorizer}\footnote{Available at \url{https://scikit-learn.org/stable/modules/generated/sklearn.feature\_extraction.text.CountVectorizer.html}, August 2022.} and \texttt{GridSearchCV}\footnote{Available at \url{https://scikit-learn.org/stable/modules/generated/sklearn.model\_selection.GridSearchCV.html}, August 2022.}, both from the Scikit-Learn Python library\footnote{Available at \url{https://scikit-learn.org}, August 2022.}, with the parameter ranges in Listing \ref{configuration_parameters}. The optimal parameter settings were \texttt{maxdf = 0.5}, \texttt{mindf = 0.01} and \texttt{ngramrange = (1,2)} (that is, the textual features selected were uniwords and biwords). Let us remark that we only used \SI{20}{\percent} of the data set in this analysis to avoid future bias in the \textsc{ml} models. Parameters \texttt{maxdf} and \texttt{mindf} respectively allow ignoring terms whose document frequencies are strictly higher (corpus-specific stop words) and lower (cut-off) than the given thresholds. Finally, \texttt{ngramrange} indicates the lower and upper boundaries for the extraction of word $n$-grams.
 
\begin{lstlisting}[frame=single,caption={Parameter ranges for the generation of $n$-grams.}, label={configuration_parameters}]
maxdf:      [0.9,0.7,0.5]
mindf:      [0.1,0.01,0.001]
ngramrange: [(1,1),(1,2),(1,3)]
\end{lstlisting}

\subsubsection{Feature analysis and selection}
\label{sec:feature_analysis_selection_results}

We estimated the contribution of the categorical features to the target using the Spearman's rank correlation coefficient (see Section \ref{sec:feature_selection_architecture}). Table \ref{tab:heatmaps} shows the features with a correlation coefficient of \SI{5}{\percent} at least. The levels endorse the appropriateness of the \textsc{ml} classification approach.

\begin{table}[!htbp]
\centering
\caption{\label{tab:heatmaps}Most correlated categorical features with the target.}
\begin{tabular}{ccc}
\toprule
\bf Feature & \bf Correlation value\\\midrule
\textit{Case type} & 0.05\\
\textit{Court} & 0.16\\
\textit{Jurisdiction} & 0.30\\\bottomrule
\end{tabular}
\end{table}

We then selected features using the transformer wrapper \texttt{SelectFromModel}\footnote{Available at \url{https://scikit-learn.org/stable/modules/generated/sklearn.feature_selection.SelectFromModel.html}, August 2022.} method from Scikit-Learn with a Random Forest Classifier and \num{20} estimators. In the end, the final set of features consisted of \textit{case type}, \textit{court}, \textit{decision}, \textit{instance type}, \textit{jurisdiction} and \textit{resolution type} plus \num{1928} out of \num{9581} uniword and biword features.

\subsection{Classification results}
\label{sec:classification_results}

We evaluated diverse \textsc{ml} algorithms based on their promising performance in similar classification problems \citep{Aljedani2021,Sengupta2021,Chen2022} and their potential for explainability \citep{BarredoArrieta2020}. The choices for both the \textsc{bts} and \textsc{mts} strategies are:

\begin{itemize}

 \item Extra Tree Classifier in both single (\textsc{etc}) and ensemble configurations (\textsc{eetc})\footnote{Available at \url{https://scikit-learn.org/stable/modules/generated/sklearn.tree.ExtraTreeClassifier.html\#sklearn.tree.ExtraTreeClassifier} and \url{https://scikit-learn.org/stable/modules/generated/sklearn.ensemble.ExtraTreesClassifier.html\#sklearn.ensemble.ExtraTreesClassifier}, August 2022.} \citep{Baby2021}
 
 \item Decision Tree (\textsc{dt})\footnote{Available at \url{https://scikit-learn.org/stable/modules/generated/sklearn.tree.DecisionTreeClassifier.html\#sklearn.tree.DecisionTreeClassifier}, August 2022.} \citep{Trabelsi2019} 
 
 \item Random Forest (\textsc{rf})\footnote{Available at \url{https://scikit-learn.org/stable/modules/generated/sklearn.ensemble.RandomForestClassifier.html\#sklearn.ensemble.RandomForestClassifier}, August 2022.} \citep{Parmar2019}
 
\end{itemize}

In this section we present the final performance evaluation of our system. Firstly, hyper-parameters were optimised with \texttt{GridSearchCV} using again \SI{20}{\percent} of the data set to avoid bias in the \textsc{ml} models. The classifiers were independently optimised for the two base strategies in Section \ref{sec:classification_architecture}.

Listing \ref{lst:hype} displays the parameter ranges for \textsc{etc} (no-ensemble), \textsc{eetc} (ensemble), \textsc{dt} (no-ensemble), and \textsc{rf} (ensemble) models. Table \ref{tab:hyperparameter} shows the selections of hyper-parameter configurations, where:

\begin{lstlisting}[frame=single,caption={Hyper-parameter configuration ranges.},label={lst:hype}]
weight:      [None,balanced]
depth:       [100, 500, None]
samplessplit:[2, 50, 100]
samplesleaf: [1, 50, 100]
criterion:   [gini, entropy]
splitter:    [best, random] #No-ensemble
estimator:   [50,100,200]   #Ensemble 

\end{lstlisting}

\begin{itemize}
 \item \textbf{Weight} represents the weights associated with the target classes, where all classes are assigned weight one if the parameter is not set. 

 \item \textbf{Depth} indicates the maximum tree depth. If not set, nodes are expanded until all leaves are pure or all leaves contain less than \texttt{samples split} samples.
 
 \item \textbf{Samples split} is the minimum number of samples required to split an internal node.
 
 \item \textbf{Samples leaf} is the minimum number of samples required at a leaf node.
 
 \item \textbf{Criterion} is the function used to measure the split quality. Supported criteria are \texttt{gini} for Gini impurity and \texttt{entropy} for information gain.
 
 \item \textbf{Splitter} indicates the strategy to choose the split at each node. Supported strategies are \texttt{best} to choose the best split and \texttt{random} to choose the best random split.
 
 \item \textbf{Estimator} is the number of trees in the model.
 \end{itemize}

\begin{table}[!htbp]
\small
\centering
\caption{\label{tab:hyperparameter} Hyper-parameters for the \textsc{ml} models.}
\begin{tabular}{lccccp{1.5cm}p{1.5cm}cc}
\toprule
\bf Model & \bf Strategy & \bf Weight & \bf Criterion & \bf Depth & \bf \centering Samples leaf & \bf \centering Samples split & \bf Splitter & \bf Estimators \\\midrule

\multirow{2}{*}{\textsc{etc}} & BTS & None & gini & 100 & \centering 10 & \centering 50 & best & -\\
 & MTS & None & gini & None &\centering 1 &\centering 100 & best & -\\
 \multirow{2}{*}{\textsc{eetc}} & BTS & balanced & entropy & 500 & \centering 1 & \centering 50 & - & 100 \\
 & MTS & None & gini & 100 & \centering 1 & \centering 2 & - & 100 \\
 \multirow{2}{*}{\textsc{dt}} & BTS & None & gini & 500 &\centering 1 &\centering 50 & best & -\\
 & MTS & None & gini & 100 &\centering 1 &\centering 50 & best & -\\
 \multirow{2}{*}{\textsc{rf}} & BTS & balanced & gini & 100 &\centering 1 &\centering 50 & - & 200 \\
 & MTS & None & gini & 100 &\centering 10 &\centering 2 & - & 200 \\ \bottomrule
\end{tabular}
\end{table}

Each classifier within each strategy in Section \ref{sec:classification_architecture} was trained and tested with 10-fold cross-validation \citep{Berrar2019}. Table \ref{tab:classifiers_results} shows the results. Overall, the \textsc{rf} model with the \textsc{mts} strategy attained the best performance, except for the recall, which was higher with this same model using the \textsc{bts} strategy. The latter took much longer to train, specially with the \textsc{rf} (1911.29 seconds) and \textsc{eetc} (1546.67 seconds) models, due to the large number of binary classifiers involved. These models seemed to fail to differentiate between classes sharing $n$-gram values, for example. The \textsc{mts} strategy achieved notable improvements, of up to \SI{10}{\percent} in exact match and precision. Accuracy was around \SI{74}{\percent} with the \textsc{rf} model using the \textsc{mts} strategy, which was the best overall solution, achieving a precision above \SI{85}{\percent}. Compared to other works in the literature (see Section \ref{sec:literature_comparison}), the performance is satisfactory.

\begin{table*}[!htbp]
\small
\centering
\caption{\label{tab:classifiers_results}Performance (\%) of the \textsc{ml} classification models.}
\begin{tabular}{ccp{1cm}ccccccccS[table-format=4.2]}
    \cmidrule(lr){1-12}
    {\bf Strategy} & {\bf Model} & {\bf \centering Exact match} & {\bf Acc.} & \multicolumn{2}{c}{\bf Precision} & \multicolumn{2}{c}{\bf Recall} & \multicolumn{2}{c}{\bf \textit{F}-measure} & {\bf HL} & {\bf Train (s)} \\
    \cmidrule(lr){5-6}
    \cmidrule(lr){7-8}
    \cmidrule(lr){9-10}
    & & & & Macro & Micro & Macro & Micro & Macro & Micro & & \\
    \midrule
\multirow{4}{*}{\textbf{BTS}} 
&\textsc{etc} &\centering 36.10 & 48.48 & 66.38 & 55.80 & 43.07 & 53.64 & 50.48 & 52.72 & 1.98 & 13.55\\
&\textsc{eetc} &\centering 45.55 & 64.96 & 71.06 & 69.82 & \bf 67.69 & \bf 79.39 & 66.74 & 71.37 & 1.65 & 1546.67\\
&\textsc{dt} &\centering 35.68 & 49.52 & 60.36 & 56.35 & 47.05 & 56.79 & 52.30 & 54.33 & 2.10 & 15.30\\
&\textsc{rf} &\centering 52.80 & 69.05 & 75.43 & 75.15 & 67.14 & 79.17 & \bf 69.79 & 74.42 & 1.38 & 1911.29\\\midrule
    
\multirow{4}{*}{\textbf{MTS}} & \textsc{etc} &\centering 44.77 & 56.05 & 58.05 & 65.18 & 44.85 & 58.62 & 49.09 & 59.92 & 2.20 & 5.79\\
& \textsc{eetc} &\centering 57.94 & 69.80 & 82.82 & 81.41 & 54.97 & 70.76 & 61.74 & 73.85 & 1.46 & 274.98\\
& \textsc{dt} &\centering 45.00 & 56.96 & 57.68 & 65.64 & 47.44 & 60.51 & 51.14 & 61.07 & 2.18 & 7.30\\
& \textsc{rf} &\centering \bf 62.32 & \bf 74.08 & \bf 86.49 & \bf 85.41 & 59.96 & 75.03 & 66.98 & \bf 78.06 & \bf 1.25 & 423.55\\   
  
\bottomrule
\end{tabular}
\end{table*}

\subsection{Explainability of the results}
\label{sec:explainability_results}

In this section we describe the automatic explanation of the decisions of the \textsc{ml} classifier by the explainability module. Algorithm \ref{alg:explainability} describes the explainability process for a decision. It traverses the branches of the decision trees of the model from their roots to the classification outcome at the leaves. The final decision is taken by the majority from the labels obtained by the decision trees. 

\begin{figure}[!htbp]
\centering
\includegraphics[width=0.75\textwidth]{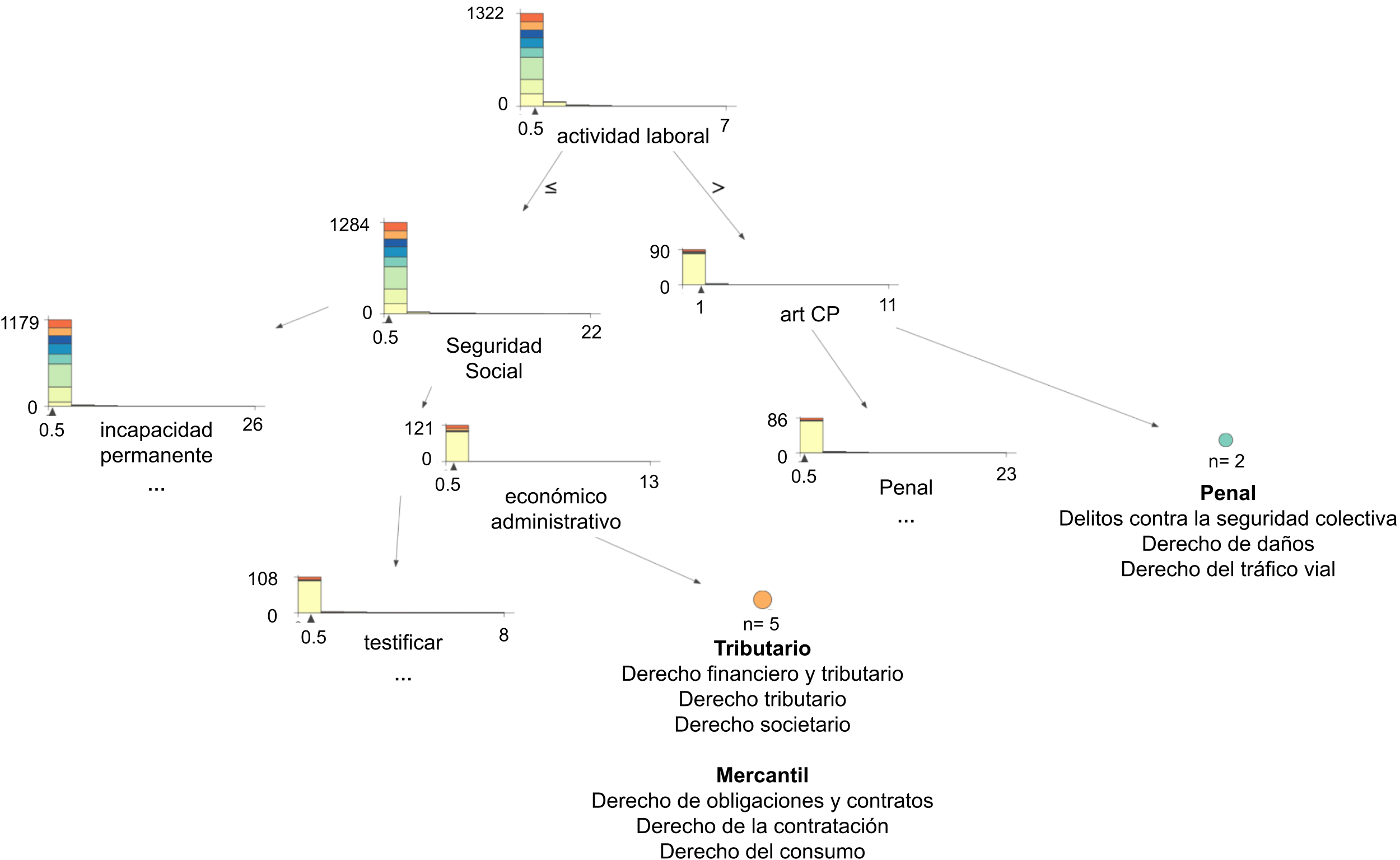}
\caption{\label{fig:partial_tree} Enhanced graph-based representation of a partial decision tree of the \textsc{rf} model.}
\end{figure}

More in detail, each estimator consists of an independent decision tree. For each tree, all $n$-grams involved in the prediction process are obtained and grouped into a dictionary, ordered by frequency. The relevance of the input features is calculated with Local Interpretable Model-Agnostic Explanations (\textsc{lime}\footnote{Available at \url{https://github.com/marcotcr/lime}, August 2022.}) \citep{Ribeiro2016}, which applies random perturbation and feature selection to determine the relevance. Only the first seven appearances in the previously mentioned ordered dictionary are taken from the \textsc{lime} results. Additionally, to enrich the explanations, the results are complemented with the values of the columns of \textit{case type}, \textit{court}, \textit{decision}, \textit{decision type}, \textit{instance type}, \textit{jurisdiction} and \textit{resolution type}. Explainability is further supported by a visual composition of the decision tree as shown in Figure \ref{fig:partial_tree}.

Particularly, Figure \ref{fig:partial_tree} shows an enhanced representation of a partial decision tree of the \textsc{rf} model, as presented by the module. It depicts the substantive order and the law categories of a judgement using the \texttt{dtreeviz} library\footnote{Available at \url{https://pypi.org/project/dtreeviz}, August 2022.}. The representation combines textual and visual information. At each node split, left branches correspond to word $n$-grams frequencies at least equal to a given threshold that is automatically set by the classifier model based on feature relevance. Right branches correspond to word $n$-grams frequencies that exceed that threshold. These thresholds are generated during the training stage of the classifier and they are automatically updated at each iteration of the model \citep{Zhang2012}. Once the training phase is finished, the thresholds are taken as a fixed reference when explaining each legal judgement at the testing stage. However, they usually change along the levels of the decision path from the root down to the leaves and they have a different effect on the classification decisions of different judgements. In the end, the number of splits corresponds to the depth of the classification decision. In this example, two predictions of the model are displayed, with one and two labels respectively: 

\begin{algorithm*}[ht!]
 \scriptsize
 \caption{Explainability process for a decision}\label{alg:explainability}
 \DontPrintSemicolon
 \KwIn{Decision path and individual data.} 
 \KwData{$node\_data$, $individual\_data$}
 \KwResult{Explanation of the classification outcome based on a decision path for the individual.}
 \Begin{
 
 $node\_data= [\{feature,threshold, left\_branch, right\_branch,type\}]$; \tcp*[h]{Data of the nodes in the decision path.}\;
 
 $individual\_data = [\{feature,feature(value)\}]$; \tcp*[h]{Data of the features the individual is composed of.}\;

 $outcome\_path=[]$; \tcp*[h]{To save the decision path of the outcome.}\;
 
 $node=node\_data[0]$; \tcp*[h]{It starts at the root.}\;

 \While{$node[type] \neq leaf$}
 {
    $feature= node[feature]$
    
	$threshold = node[threshold]$
	
    $right\_branch = node[right\_branch]$
    
    $left\_branch = node[left\_branch]$

    \uIf{$individual\_data[feature(value)] \leq threshold$} {
        $outcome\_path.append(feature,individual\_data[feature(value)],\mbox{``}less\mbox{''})$
        
	   	$node=node\_data[left\_branch]$
    }
    \uElse {
        $outcome\_path.append(feature,individual\_data[feature(value)],\mbox{``}more\mbox{''})$
        
	   	$node=node\_data[right\_branch]$
    }

 }
}
\KwRet{$outcome\_path$} \tcp*[h]{Returns the data for the explanation template.}\;
\end{algorithm*}

If the frequency of ``actividad laboral'' (\textit{work activity}) is less than \num{0.5} and the frequencies of ``seguridad social'' (\textit{social security}) and ``econ\'omico administrativo'' (\textit{administrative economics}) exceed \num{0.5}, the individual is assigned the following two classes:

\begin{enumerate}
    \item Substantive order: \textit{tributary}. Law categories: \textit{financial/tax law}, \textit{tax law} and \textit{corporate law}.
    \item Substantive order: \textit{mercantile}. Law categories: \textit{obligations/contracts}, \textit{contract law} and \textit{consumer law}.
\end{enumerate}

Otherwise, if the frequencies of ``actividad laboral'' (\textit{work activity}) and ``art CP'' (standing for \textit{penal code article}) exceed \num{0.5} and \num{1.0} respectively, the legal judgement is classified as:

\begin{itemize}
    \item Substantive order: \textit{penal}. Law categories: \textit{crimes against collective security}, \textit{tort law} and \textit{road traffic law}.
\end{itemize}

Classification descriptions in natural language are based on templates. Listing \ref{lst:explainability_nl} provides an example of classification description. The confidence levels characterising the classification outcome were obtained with \texttt{predict\_proba}\footnote{Available at \url{https://scikit-learn.org/0.17/modules/generated/sklearn.ensemble.RandomForestClassifier.html}, August 2022.}, based on mean predicted class probabilities of the trees in the forest.

\begin{figure}[!htbp]
\begin{lstlisting}[caption={Natural language explanation obtained from the \textsc{rf} model.},label={lst:explainability_nl}]
For sample 10 the features' values and model decision are:

- Case type: recurso de suplicación
- Court: Tribunal Superior de Justicia
- Decision: desestimatorio
- Decision type: sustantivo
- Instance type: segunda
- Jurisdiction: social
- Resolution type: sentencia

- Substantive order: social
- Law categories: derecho del trabajo, derecho de la contratacion laboral y derecho relativo al contrato de trabajo

This decision has a confidence of 88%.

The most representative terms (ngrams) and their relevance are:
- Estatuto Trabajadores -- 0.076
- recurso suplicación -- 0.070
- Jurisdicción Social -- 0.067
- suplicación -- 0.064
- trabajadores -- 0.051
- Estatuto -- 0.033

\end{lstlisting}
\end{figure}

Both graph-based and natural language explanations provided exploit the fact that the underlying model is intrinsically interpretable. This contributes to the classifier reasoning transparency and makes it more understandable to end users.

\subsection{Comparison with prior work}
\label{sec:literature_comparison}

None of the following works on the classification of legal judgements considered the automatic explainability of their decisions, and therefore this requirement did not restrict their selections of methods. We indicate in parentheses the comparison with our performance metrics when it was favourable to us. Otherwise, we explain the fundamental differences with our work. We exclude from this comparison interesting research on multi-label classification outside the domain of this work \citep{Aljedani2021,Sengupta2021,Caled2022,Teisseyre2021,Caled2019,Chen2016}, and other research that did actually deal with single-label problems \citep{Bambroo2021,Chen2022}.

One of the most relevant competing works is that by \cite{Qiu2020} on multi-label classification of judgements using convolutional neural networks. They reported excellent results (\SI{98.07}{\percent} precision, \SI{99.61}{\percent} recall and \SI{98.84}{\percent} \textit{F}-measure, all computed as macro averages) but the experimental data set was tiny (143 legal documents). It is worth mentioning that the training times were too long (approximately 3 hours).

\cite{Csanyi2022}, which addressed the same problem with a binary transformation approach, attained \SI{31}{\percent} exact match (\SI{-31}{\percent}), \SI{57}{\percent} accuracy (\SI{-17}{\percent}) and \SI{80}{\percent} \textit{F}-measure.

Finally, \cite{Song2022} presented a multi-label classification system for judgements based on deep learning. Their results were \SI{87.80}{\percent} micro precision; \SI{80.30}{\percent} micro recall and \textit{F}-measure; \SI{31.40}{\percent} macro precision (\SI{-55}{\percent}), \SI{28.70}{\percent} macro recall (\SI{-31}{\percent}) and \SI{28.40}{\percent} macro \textit{F}-measure (\SI{-39}{\percent}).
The training time was 3 hours approximately.

\section{Conclusions}
\label{sec:conclusions}

\textsc{ml} models have not been exploited in the legal domain to their maximum potential. Some users may be reluctant to accept them owing to the opacity of their decisions. 
Therefore, we propose a solution to address the complex multi-label multi-class categorisation of legal judgements and the automatic explanation of this categorisation. Our work contributes to the state of the art in legal \textsc{ai} with a hybrid system that combines \textsc{ml} for prediction and visual and natural language representation down to feature level for explanations. Thus, we both describe the result of the process and the process itself. We include an in-depth entity detection scheme to extract relevant judicial entities such as case type, court and jurisdiction among others. Anonymisation guarantees confidentiality and avoids bias in the \textsc{ml} model. We remark that this is the first study centred on Spanish, the fourth language in the world with over 500 million speakers.

Our system has attained competitive results (\SI{85}{\percent} micro precision) on a data set that has been annotated by lawyers, which endorses its interest in relieve human experts from monotonous labour-intensive classification tasks. As explained in Section \ref{sec:literature_comparison}, it is competitive with previous work without explainability capabilities.

In future work, we will consider classification based on the eXtreme Multi-label Learning (\textsc{xml}) representation of the whole Spanish legal system, and study the dependencies among the law categories. We also plan to explore language models based on transformers to train our system due to the highly specialised target domain.



\section*{Declaration of competing interest}

The authors declare no competing financial interests or personal relationships that could influence the work reported in this paper.

\section*{Acknowledgements}

This work was supported by E4Legal Analytics \textsc{sl}, and partially supported by Xunta de Galicia grants ED481B-2021-118 and ED481B-2022-093, Spain.

\bibliography{bibliography}

\end{document}